\documentclass[conference]{IEEEtran}
\IEEEoverridecommandlockouts

\RequirePackage[absolute]{textpos}
\DeclareRobustCommand*{\copyrightnote}{%
  \begin{textblock}{85}(17.5,256.75)
      \scriptsize{\noindent \copyright 2022 IEEE. Personal use of this material is permitted. Permission from IEEE must be obtained for all other uses, in any current or future media, including reprinting/republishing this material for advertising or promotional purposes, creating new collective works, for resale or redistribution to servers or lists, or reuse of any copyrighted
component of this work in other works.}%
  \end{textblock}%
    }

\usepackage{amsmath,amssymb,amsfonts}
\usepackage{algorithmic}
\usepackage{graphicx}
\usepackage{textcomp}
\usepackage{xcolor}
\def\BibTeX{{\rm B\kern-.05em{\sc i\kern-.025em b}\kern-.08em
    T\kern-.1667em\lower.7ex\hbox{E}\kern-.125emX}}
    
\usepackage[hidelinks]{hyperref}
\usepackage{placeins}

\usepackage{changes}
\usepackage[nameinlink, capitalize]{cleveref}

\usepackage[giveninits=true,sorting=none,style=ieee]{biblatex}

\newcommand{\T}{\mathrm{T}}

\newcommand{\ent}[1]{\pmb{e}_{#1}}

\newcommand{\stime}[1]{\pmb{t}_{#1}}

\newcommand{\rel}[1]{\pmb{R}_{#1}}
\newcommand{\vrel}[1]{\pmb{r}_{#1}}
\newcommand{\score}[3]{\theta_{#1,#2,#3}}
\newcommand{\sscore}[3]{\vartheta_{#1,#2,#3}}

\newcommand{\heaviside}[1]{\theta\left( #1 \right)}
\newcommand{\taus}{\tau_\mathrm{s}}
\newcommand{\taur}{\tau_\mathrm{ref}}
\newcommand{\taum}{\tau_\mathrm{m}}
\newcommand{\thresh}{u_\mathrm{th}}
\newcommand{\nlifsum}[1]{\sum_{t_{#1} \leq t^*} W_{s,i#1}}
\newcommand{\nlifsumi}[1]{\sum_{t_{#1} \leq t_{s,i}} W_{s,i#1}}

\newcommand{\citep}[1]{\cite{#1}}
\newcommand{\citet}[1]{\cite{#1}}

\usepackage{nameref}

\begin{document}

\title{Relational representation learning with spike trains}

\author{\IEEEauthorblockN{Dominik Dold}
\IEEEauthorblockA{\textit{European Space Agency, European Space Research and Technology Centre, Advanced Concepts Team}\\
Noordwijk, the Netherlands \\
}
}

\maketitle
\copyrightnote
\thispagestyle{plain}
\pagestyle{plain}
\begin{abstract}
Relational representation learning has lately received an increase in interest due to its flexibility in modeling a variety of systems like interacting particles, materials and industrial projects for, e.g., the design of spacecraft.
A prominent method for dealing with relational data are knowledge graph embedding algorithms, where entities and relations of a knowledge graph are mapped to a low-dimensional vector space while preserving its semantic structure.
Recently, a graph embedding method has been proposed that maps graph elements to the temporal domain of spiking neural networks.
However, it relies on encoding graph elements through populations of neurons that only spike once.
Here, we present a model that allows us to learn spike train-based embeddings of knowledge graphs, requiring only one neuron per graph element by fully utilizing the temporal domain of spike patterns.
This coding scheme can be implemented with arbitrary spiking neuron models as long as gradients with respect to spike times can be calculated, which we demonstrate for the integrate-and-fire neuron model.
In general, the presented results show how relational knowledge can be integrated into spike-based systems, opening up the possibility of merging event-based computing and relational data to build powerful and energy efficient artificial intelligence applications and reasoning systems.
\end{abstract}

\section{Introduction}\label{sec:introduction}
Recently, spiking neural networks (SNNs) have started to close the performance gap to their artificial counterparts commonly used in deep learning \citep{neftci2019surrogate,yin2020effective,zenke2021visualizing}.
However, it remains an open question how the temporal domain of spikes can be fully utilized to efficiently encode information -- both in biology as well as in engineering applications \citep{davies2019benchmarks,zenke2021visualizing}.
Although a lot of progress has been made when it comes to encoding and processing sensory information (e.g., visual and auditory) with SNNs \citep{zenke2018superspike,mostafa2017supervised,comsa2019temporal,kheradpisheh2019s4nn,goltz2021fast,wunderlich2021event}, there has been almost no work on applying SNNs to relational data and symbolic reasoning tasks \citep{crawford2016biologically,dold2021spikeembed}.
Here, we present a framework that allows us to learn semantically meaningful spike train representations of abstract concepts and their relationships with each other, opening up SNNs to the rich world of relational and symbolic data.

\begin{figure*}[h]
    \centering
    \includegraphics[width=0.8\textwidth, page=1]{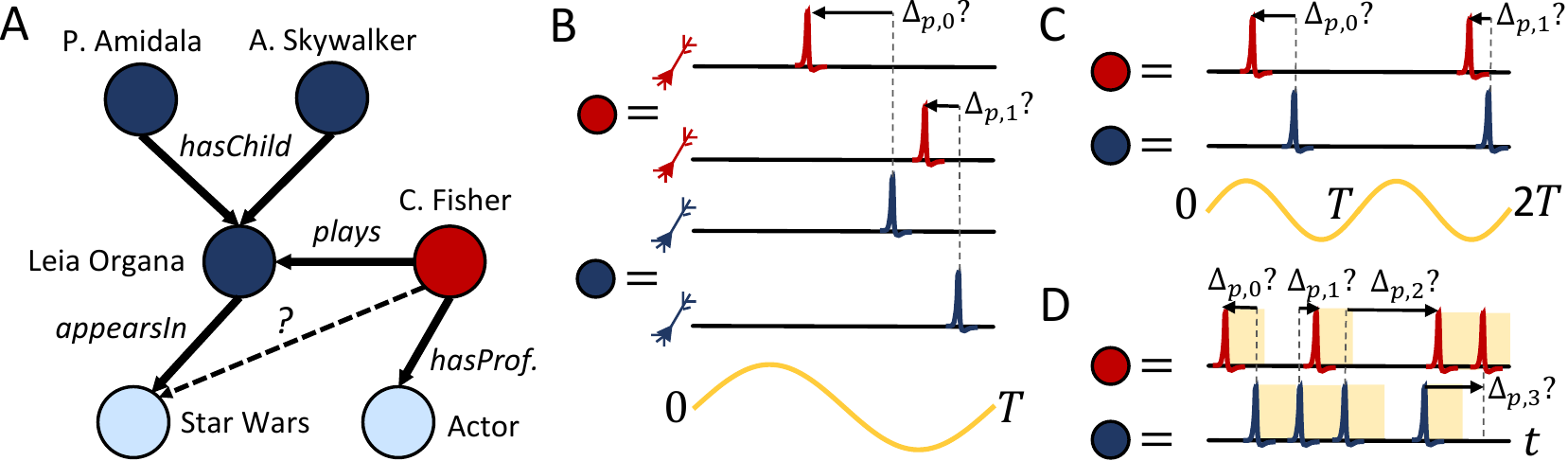}
	\caption{
	\textbf{(A)} Schematic illustration of a KG consisting of nodes (circles) and edges (lines) of different types.
	Inference on the KG consists of evaluating the plausibility of novel links (dashed line) given the information contained in the KG.
	\textbf{(B)} In SpikE, nodes are represented by the TTFS set of a population of neurons.
	Alternatively, the spike times can be coupled to the phase of a global background oscillation (orange).
	Relations are represented as spike time (or phase) differences.
	\textbf{(C)} The SpikE encoding can be reduced to spike trains by concatenating spiking periods.
	\textbf{(D)} The proposed model enables the learning of spike train embeddings with rich temporal structure, featuring refractory periods (orange) as well as variable interspike intervals and spike train lengths.
	}\vspace{-4mm}	
	\label{fig:intro}
\end{figure*}

We approach this problem from the perspective of knowledge graphs (KGs).
KGs are capable of integrating information from different domains and modalities into a unified knowledge structure \citep{auer2007dbpedia,bollacker2008freebase,singhal2012introducing}. 
In general, a KG $\mathcal{KG}$ consists of nodes representing entities $s \in \mathcal E$ and edges representing relations $p \in \mathcal R$ between these entities (\cref{fig:intro}A).
It can be summarized using semantically meaningful statements (called triples) that can be represented as \textit{(node, typed edge, node)} in graph format or \textit{(subject, predicate, object)} in human-readable form\footnote{Thus, a KG can be mathematically summarized as $\mathcal{KG} \subset \mathcal{E} \times \mathcal{R} \times \mathcal{E}$.}, for instance, \textit{(C. Fisher, plays, Leia Organa)} and \textit{(Leia Organa, appears in, Star Wars)}.
Reasoning on the KG is concerned with discovering new knowledge, e.g., evaluating whether previously unseen facts are true or false, like \textit{(C. Fisher, appears in, Star Wars)}.
This is generally known as the link prediction or KG completion task. 

A widely adopted approach for such tasks is graph embedding \cite{nickel2015review,hamilton2017representation,ruffinelli2019you}, where elements of the graph are mapped into a low-dimensional vector space while conserving certain graph properties.
This approach has recently been adapted to SNNs \citep{dold2021spikeembed} and extended to spike-based relational graph neural networks \citep{chian2021learning}.
However, in \citep{dold2021spikeembed}, each element of the graph is represented by a population of neurons where every neuron only spikes once -- a somewhat inefficient approach considering that spike time patterns are very expressive.
To address this point, we present an extension of the spike-based embedding approach of \citep{dold2021spikeembed} where nodes are represented by the spike train of single neurons -- eliminating the need of population codes for representing graph elements by fully harnessing the temporal domain of spike trains.

In the following, we first briefly review the single-spike graph embedding model presented in \citep{dold2021spikeembed} before extending it to spike trains in \cref{sec:current}. 
The introduced coding scheme can be implemented with arbitrary spiking neuron models trainable via gradient descent, which we demonstrate as a proof of concept for integrate-and-fire neurons in \cref{sec:neuronal}.
Finally, we conclude with a brief discussion of the presented results and future prospects in \cref{sec:discussion}.

\section{Single-spike embeddings}\label{sec:previous}
Recently, a coding scheme referred to as SpikE for embedding elements of a knowledge graph into the temporal domain of spiking neurons has been proposed \citep{dold2021spikeembed,chian2021learning}.
In SpikE, knowledge graph entities (nodes) are represented as spike times of neuron populations, while relations (edge types) are represented as spike time differences between populations.
More specifically, a node $s$ in the graph is represented by the set of time-to-first-spike (TTFS) values $\stime{s} \in \mathbb{R}^N$ of a population of $N \in \mathbb{N}$ spiking neurons.
Relations are encoded by a $N$-dimensional vector of spike time differences $\pmb{\Delta}_p \in \mathbb{R}^N$.
The plausibility of a triple $(s,p,o)$ is evaluated based on the discrepancy between the spike time differences of the node embeddings, $\stime{s} - \stime{o}$, and the relation embedding $\pmb{\Delta}_p$ (\cref{fig:intro}B)
\begin{equation}
    \sscore{s}{p}{o} = -\sum_j \big\|\stime{s} - \stime{o} - \pmb{\Delta}_p \big\|_j \,, \label{eq:asymm}
\end{equation}
with $\|\cdot\|$ denoting the absolute value.
Here, both the order and the difference of the spike times are important, which allows for modeling asymmetric relations in a KG.
Alternatively, the scoring function $\sscore{s}{p}{o}$ can also be defined symmetrically
\begin{equation}
    \sscore{s}{p}{o} = -\sum_j \big\| \|\stime{s} - \stime{o}\| - \pmb{\Delta}_p \big\|_j \,. \label{eq:symm}
\end{equation}
i.e., where only the absolute spike time difference is taken into account, which only allows for modeling symmetric relations in a KG\footnote{Which scoring function to use strongly depends on the underlying structure of the KG. As a default, we use the asymmetric scoring function.}.
If a triple is valid, then the patterns of node and relation embeddings match, leading to $\sscore{s}{p}{o} \approx 0$, i.e., $\stime{s} \approx \stime{o} + \pmb{\Delta}_p$.
If the triple is not valid, we have $\sscore{s}{p}{o} < 0$, with lower values representing lower plausibility scores.
Given a KG, such spike embeddings are found using gradient descent-based optimization with the objective of assigning high scores to the existing edges in the training KG.

Since the embeddings are represented by single spike times of neuron populations in a fixed (and potentially repeating) time interval, the used coding scheme is identical to phase-based coding, meaning that every neuron spikes at a certain phase of a repeating background oscillation (\cref{fig:intro}B).
Relations can be identified as spike times as well: in case of a symmetric scoring function, all relation embeddings are strictly positive and can be encoded as the TTFS set of a population of neurons.
In case of the asymmetric scoring function, both the absolute value as well as the order of spike times (the sign of their difference) have to be represented, which could, for instance, be done by encoding relations via two populations of spiking neurons -- one population used for positive signs and one for negative signs\footnote{To clarify, in this case $\pmb{\Delta}_p$ would be represented by $\pmb{\Delta}_p^+$ and $\pmb{\Delta}_p^-$. If $\Delta_{p,i} < 0$, then $\Delta_{p,i}^- = \|\Delta_{p,i} \|$ and the $i$-th neuron in $\pmb{\Delta}_p^+$ does not spike.}. 

\section{Spike train embeddings}\label{sec:current}
\begin{figure*}[t!]
    \centering
    \includegraphics[width=0.85\textwidth]{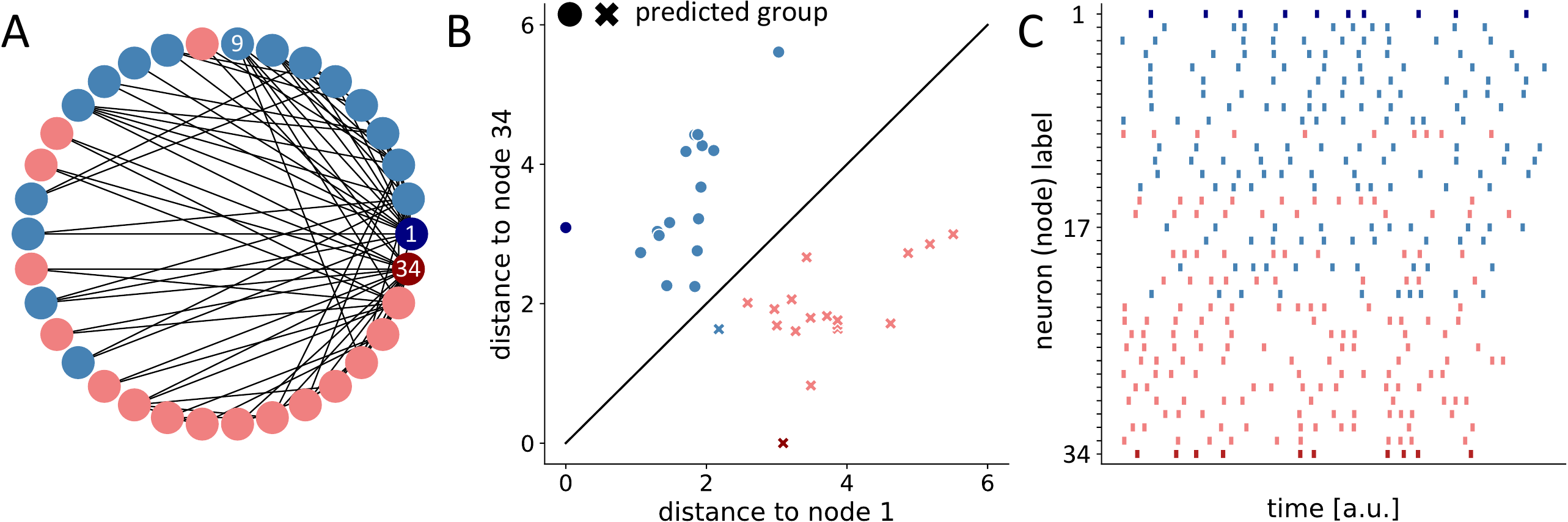}
	\caption{
	\textbf{(A)} Zachary Karate Club graph, consisting of the 34 members (circles) with edges indicating which members interacted with each other outside of the Karate club.
	The Karate club splits into two groups marked by red and blue.
	\textbf{(B)} Clustering of the Karate club members using spike train embeddings that were learned from the connectivity of the graph shown in (A).
	\textbf{(C)} Learned spike train embedding for all members, with one neuron per member (node).
	}\vspace{-4mm}	
	\label{fig:zachary}
\end{figure*}

In principle, the SpikE representation can be extended to spike trains: instead of interpreting the spike embedding vector as a population of neurons that spike in a given time interval $[0,T]$, we can interpret it as the spike times of a single neuron that only spikes once per consecutive time interval $[0,T]$, $[T, 2T]$, ..., $[(N-1)T, N T]$ (\cref{fig:intro}C).
However, the resulting spike trains are rather rigid and lack features like tonic bursting, refractory periods or very long interspike intervals.
Furthermore, spike times are artificially separated into consecutive time intervals, which means that spikes cannot occur outside of their assigned time intervals, and therefore all obtained spike train embeddings will span the same total time period.
In the following, we present a model that does not suffer from these downsides and allows learning of graph embeddings that can be plausibly interpreted as spike trains with rich and complex temporal structure.

\subsection{Abstract model}

We first introduce an abstract model for learning spike train embeddings of graph elements, which will be mapped to an actual spiking neuron model in \cref{sec:neuronal}.
Interpreting a node embedding vector $\stime{s} \in \mathbb{R}^{N}$ as a vector of consecutive spike times of a single neuron requires that the elements of the vector are ordered ($\forall i$, $t_{s,i} > t_{s,i+1}$), which has to be true at all times for all embeddings while training on a KG.

A spike train representation satisfying this requirement can be constructed as follows.
First, for each node $s$ we define a vector of interspike intervals $\pmb{I}_s$ given by 
\begin{equation}
    I_{s,i} = \|\tilde{I}_{s,i}\| \,,
\end{equation}
with $\tilde{I}_{s,i} \in \mathbb{R}$.
We then define the spike times of the neuron encoding node $s$ as
\begin{align}
    t_{s,i} &= \frac{1}{Z} \sum_{j\leq i} I_{s,j}  \,, 
\end{align}
where $Z = \sqrt{\sum_{k=0}^N I_{s,k}^2}$ is a normalizing term\footnote{In the graph embedding model TransE \citep{bordes2013translating} the embedding vector is also normalized in order to prevent vector components from growing too large. When implementing the proposed coding with spiking neuron models, one can set $Z=1$ as the possible spike times are limited by the neuron's inputs.}.
The resulting embedding vector $\stime{s}$ has the required property that all entries remain ordered during training, hence we can interpret each entry as an actual spike time and the whole vector as a spike train of fixed length.
Note that changing an interspike interval affects all subsequent spike times, similar to how changing the presynaptic input of a neuron to alter its spike behavior at a specific time can also influence its later spiking behavior.

Relations are encoded as $N$-dimensional vectors $\pmb{\Delta}_p \in \mathbb{R}^N$.
Each entry $\Delta_{p,i}$ represents the expected spike time difference between the $i$-th spike of two node embedding neurons.
In other words, if two entities $s$ and $o$ are related via $p$, then $\stime{s} - \stime{o} \approx \pmb{\Delta}_p$ (\cref{fig:intro}D).
Similar to SpikE, a link in the knowledge graph can be scored either via an asymmetric scoring function (\cref{eq:asymm}) or a symmetric one (\cref{eq:symm}).

To learn the embeddings, we optimize a margin-based loss function $\mathcal{L}$ via gradient descent
\begin{equation}
    \mathcal{L} = \frac{1}{|\mathcal{D}|} \sum_{(s,p,o) \in \mathcal{D}}\left[\gamma - \sscore{s}{p}{o} + \sscore{s'}{p}{o'}\right]_+ \,, \label{eq:loss}
\end{equation}
where in $\sscore{s'}{p}{o'}$ either $s$ or $o$ is replaced with a random entity and $\left[x\right]_+ = \mathrm{max}\left(0,x\right)$.
$\gamma$ is a scalar hyperparameter and $|\mathcal{D}|$ is the number of triples contained in the training graph $\mathcal{D}$.
During training, we update both $\pmb{\tilde{I}}_{s}$ to change the spike times as well as the relation embeddings $\pmb{\Delta}_p$, $\forall s \in \mathcal{E}$ and $\forall p \in \mathcal{R}$.

\subsection{Refractory periods}

In the aforementioned representation, consecutive spikes of a spike train can be coincident -- something that is not observed in actual spiking neuron models.
This can be alleviated by introducing an absolute refractory period $\taur$,
\begin{align}
    t^\mathrm{ref}_{s,i} &= \frac{1}{Z} \sum_{j\leq i} I_{s,j} + i \cdot \taur  \,, 
\end{align}
which leaves scores invariant since $t^\mathrm{ref}_{s,i}-t^\mathrm{ref}_{o,i} = t_{s,i} - t_{o,i}$.
\subsection{Experiments}\label{sec:results}
\subsubsection{Community prediction}\label{sec:detection}

We first illustrate how the proposed spike train embedding model works with a simple and well-known community prediction task: the Zachary Karate Club \citep{zachary1977information}.
The Zachary Karate Club data is a social graph of 34 members of a Karate club that split into two groups (\cref{fig:zachary}A, groups are marked with different colors).
It summarizes how the members interacted with each other outside of the Karate club, with edges representing past interactions.
The task is to predict -- given the structure of this graph -- which group a member will join, with the heads of the groups being member 1 and member 34.

We can solve this task by learning a spike train embedding for each member and then using the scoring functions $\sscore{s}{\mathrm{interact}}{1}$ and $\sscore{s}{\mathrm{interact}}{34}$ to predict group membership -- i.e., if  $\sscore{s}{\mathrm{interact}}{1} > \sscore{s}{\mathrm{interact}}{34}$, member $s$ will join the group of member 1 and if $\sscore{s}{\mathrm{interact}}{34} > \sscore{s}{\mathrm{interact}}{1}$, they will join the group of member 34.
Since there is only one relation type, we fix the relation embedding $\pmb{\Delta}_\mathrm{interact} = 0$.
This way, the graph embedding task becomes a pure clustering task, where members of the same group will be represented by similar spike trains.

As shown in \cref{fig:zachary}B, the decision which group a member joins can be predicted from the learned spike train embeddings, with the exception of member 9, whose decision has also been falsely predicted in the original studies on the Zachary Karate Club \citep{zachary1977information}.
Moreover, the learned spike trains (\cref{fig:zachary}C) contain complex temporal features like small tonic bursts (such as neuron 19) or large interspike intervals (such as neuron 3) to encode how related members are given their past interactions.

\subsubsection{Link prediction}\label{sec:linkprediction}

To evaluate the model on general KGs we employ the following two metrics from the graph embedding literature: the mean reciprocal rank (MRR) and the hits@k metric \citep{ruffinelli2019you}.
The idea behind both metrics is to evaluate how well a test triple $(s,p,o)$ is ranked compared to other, possibly implausible, triples $(s,p,o') \notin \mathcal{KG}$ or $(s',p,o) \notin \mathcal{KG}$, where $s' \in \mathcal{E} \setminus s$ and $o' \in \mathcal{E} \setminus o$.
We expect that a good model assigns high ranks to test triples, meaning that they are scored higher than alternative links with unknown plausibility.

To obtain both metrics, we use our learned embeddings to first evaluate the score $\sscore{s}{p}{o}$ of a test triple $(s,p,o)$.
In addition, we also calculate the score for every alternative fact $(s,p,o') \notin \mathcal{KG}$, where $o'$ are all suitable entities contained in the graph\footnote{We use so-called filtered metrics here, i.e., entities $s'$ (or $o'$) that lead to triples $(s',p,o)$ (or $(s,p,o')$) that are known to be valid (e.g., from the training graph) are not considered.}.
With this, we can create a list containing the test triple and corresponding alternative links, sorted in descending order of scores
\begin{align}
    \mathcal{O}^{spo} = \left[(s,p,o')\, |\, o' \in \mathcal{E}, (s,p,o') \notin \mathcal{KG}\setminus (s,p,o)\right]\,,
\end{align}
with $\vartheta_{\mathcal{O}^{spo}_i} > \vartheta_{\mathcal{O}^{spo}_{i+1}}$ $\forall i$.
The rank of a test triple is given by its position in this list (i.e., the triple with the highest score has a rank of 1), and the reciprocal rank (RR) is the inverse of the rank
\begin{equation}
    \mathrm{RR}^{spo}_\mathcal{O} = \frac{1}{j} \quad \mathrm{with} \quad j \quad \text{such that} \quad \mathcal{O}^{spo}_j = (s,p,o) \,.
\end{equation}
For hits@k, we measure whether the test triple is among the k highest-scored triples in $\mathcal{O}^{spo}$,
\begin{equation}
    h^{spo}_{\mathcal{O},k}=\begin{cases}
    1, & \text{if $(s,p,o) \in (\mathcal{O}^{spo}_1,\ ..., \mathcal{O}^{spo}_k)$}.\\
    0, & \text{otherwise}.
  \end{cases}
\end{equation}
The same process is repeated for replacing the subject, resulting in another sorted list
\begin{align}
    \mathcal{S}^{spo} = \left[(s',p,o)\, |\, s' \in \mathcal{E}, (s',p,o) \notin \mathcal{KG}\setminus (s,p,o)\right]\,,
\end{align}
with $\vartheta_{\mathcal{S}^{spo}_i} > \vartheta_{\mathcal{S}^{spo}_{i+1}}$ $\forall i$, from which the corresponding reciprocal rank ($\mathrm{RR}^{spo}_\mathcal{S}$) and value for hits@k ($h^{spo}_{\mathcal{S},k}$) can be extracted.
The MRR and hits@k metrics are obtained by averaging over all test triples $\mathcal{T}$
\begin{align}
    \mathrm{MRR} &= \frac{1}{2 |\mathcal{T}|} \sum_{(s,p,o) \in \mathcal{T}} \sum_{i \in \{ \mathcal{S}, \mathcal{O}\}} \mathrm{RR}^{spo}_i  \,, \\
    \mathrm{hits@}k &= \frac{1}{2 |\mathcal{T}|} \sum_{(s,p,o) \in \mathcal{T}} \sum_{i \in \{ \mathcal{S}, \mathcal{O}\}} h^{spo}_{i,k} \,.
\end{align}
Thus, MRR measures the average reciprocal rank of test triples, while hits@k measures how often, on average, the test triple is among the k highest-scored triples.
Both metrics take values between 0 and 1, with 1 being the best achievable value.

The model is evaluated on the following data sets:
\begin{itemize}
    \item \textbf{FB15k-237} \citep{toutanova2015observed}: a KG derived from Freebase that is widely used to benchmark graph embedding algorithms.
    \item \textbf{CoDEx-S} \citep{safavi2020codex}: a KG constructed for link prediction tasks from Wikidata and Wikipedia.
    \item \textbf{IAD} \citep{soler2021graph,dold2021spikeembed}: a KG modeling the components and their interactions in an Industrial Automation Demonstrator.
    \item \textbf{UMLS} \citep{mccray2003upper}: a biomedical KG holding facts about diseases, medications and chemical compounds.
    \item \textbf{Kinships} \citep{kemp2006learning}: a social KG containing the family relationships between people (mother, uncle, sister, etc.).
\end{itemize}
The data set statistics are summarized in \cref{tab:datasets}.
\begin{table}[hb!]
\caption{Statistics of the data sets used in the experiments.}
\label{tab:datasets}
\center
\begin{tabular}{ c|  c|  c | c}
 Data set & \#Entities & \#Relations & \#Triples \\
 \hline \hline
FB15k-237 & 14,541 & 237 & 310,116 \\
CoDEx-S & 2,034 & 42 & 36,543\\
IAD  & 3,529 & 39 & 12,399\\
UMLS & 135 & 49 & 5,216\\
Kinships & 104 & 25 & 10,686\\
\hline\hline
\end{tabular}
\end{table}
We compare our results with those obtained using
\begin{itemize}
    \item \textbf{TransE} \citep{bordes2013translating}: nodes and relations are both embedded as vectors $\ent{s}$ and $\vrel{p}$, respectively.
    A triple is scored using $\sscore{s}{p}{o}^\mathrm{TransE} = -\sum_j \big\|\ent{s} - \ent{o} - \vrel{p} \big\|_j$, with values closer to 0 representing better scores.
    \item \textbf{RESCAL} \citep{nickel2011three}: nodes are embedded as vectors $\ent{s}$ and relations as matrices $\rel{p}$. A triple is scored using the product $\sscore{s}{p}{o}^\mathrm{RESCAL} = \ent{s}^\T \rel{p} \ent{o}$, with higher scores being better.
\end{itemize}

Comparing the results with TransE is especially interesting, since our proposed model is an extension of it and therefore faces similar limitations.
In contrast, RESCAL is based on tensor factorization and allows the simultaneous modeling of symmetric and asymmetric relations, which is only possible to a limited extent with our model or with TransE.

\begin{figure}[t]
    \centering
    \includegraphics[width=\columnwidth, page=2]{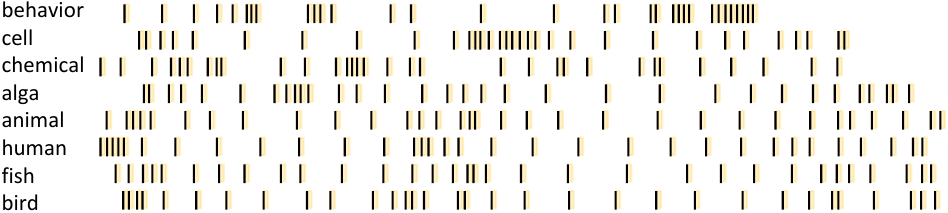}
	\caption{Spike train embeddings of concepts of the UMLS data set. Refractory periods are marked in yellow.
	}\vspace{-4mm}	
	\label{fig:UMLSpikes}
\end{figure}

On all data sets, our model (here abbreviated as SpikTE) achieves comparable results to TransE (\cref{tab:mrr}).
Since the Kinships data set contains many symmetric relations, we used a symmetric scoring function for it.
RESCAL reaches similar performance to both SpikTE and TransE, and only strongly outperforms them on the Kinships data set that contains a large amount of exclusively symmetric or asymmetric relations.
The values obtained for our references are consistent with results reported in the literature \citep{nathani2019learning,dold2021spikeembed,dold2021tensor,chian2021learning}, but can be further improved with more involved optimization schemes \citep{ruffinelli2019you}. %
\begin{table}[b!]
\caption{Test performance on link prediction tasks.}
\label{tab:mrr}
\begin{center}\renewcommand{\arraystretch}{1.2}
\begin{tabular}{c | c | c c c}
\hline\hline
Data set & Model & MRR & hits@1 & hits@3 \\
\hline\hline
FB15k-237
 & SpikTE (our)  & 0.21 & 0.13 & 0.23 \\
 & TransE  & 0.21 & 0.12 & 0.24 \\
 & RESCAL  & 0.28 & 0.20 & 0.31 \\
\hline 
CoDEx-S
 & SpikTE  &  0.30 & 0.18 & 0.34\\
 & TransE & 0.35 & 0.21 & 0.41 \\
 & RESCAL &  0.40 & 0.29 & 0.45\\
\hline
IAD 
 & SpikTE   & 0.66 & 0.63 & 0.68 \\
 & TransE   & 0.66 & 0.62 & 0.67 \\
 & RESCAL   & 0.61 & 0.58 & 0.62 \\
\hline 
UMLS
 & SpikTE   & 0.81 & 0.68 & 0.94 \\
 & TransE   & 0.81 & 0.68 & 0.93 \\
 & RESCAL   & 0.88 & 0.79 & 0.97 \\
\hline
Kinships
 & SpikTE (sym.) & 0.47 & 0.30 & 0.54 \\
 & TransE (sym.)  & 0.48 &  0.31 & 0.54 \\
 & RESCAL   & 0.81 & 0.71 & 0.90 \\
\hline \hline
\end{tabular}
\label{tab1}
\end{center}
\end{table}
To illustrate how inference with the learned embeddings works, we can ask questions as incomplete triples $(s,p,?)$ and identify the entities that are ranked best.
We show this here for the UMLS data set that has a high degree of interpretability.
For example, the query \textit{(bird, is a, ?)} yields the following proposals (sorted by descending rank): \textit{entity}, \textit{physical object}, \textit{organism}, \textit{vertebrate}, \textit{animal}, \textit{invertebrate} and \textit{anatomical structure}.
According to our model, the least likely solutions are (sorted by descending rank) \textit{language}, \textit{research device}, \textit{medical device}, \textit{regulation or law} and \textit{clinical drug}.
Spike trains encoding abstract concepts like \textit{bird} are illustrated in \cref{fig:UMLSpikes}.

\subsubsection{Unequal number of spikes}

The proposed coding scheme requires that all spike trains have the same length.
In real spiking systems this cannot always be guaranteed.
Thus, we further investigate how varying the spike train length affects the performance of the proposed embedding model on the UMLS link prediction task. 

To achieve this, we choose the number of spikes for every embedding neuron randomly from a normal distribution.
When calculating the score of a triple $\sscore{s}{p}{o}$ (e.g., during training or test time), we then only sum over the matching spike times of $\stime{s}$ and $\stime{o}$, i.e., if $\stime{s} \in \mathbb{R}^n$, $\stime{o} \in \mathbb{R}^m$ and $n<m$, then $\sscore{s}{p}{o} = -\sum_{j \leq n} \big\|\stime{s} - \stime{o} - \pmb{\Delta}_p \big\|_j\,$.

\begin{figure}[t]
    \centering
    \includegraphics[width=0.98\columnwidth]{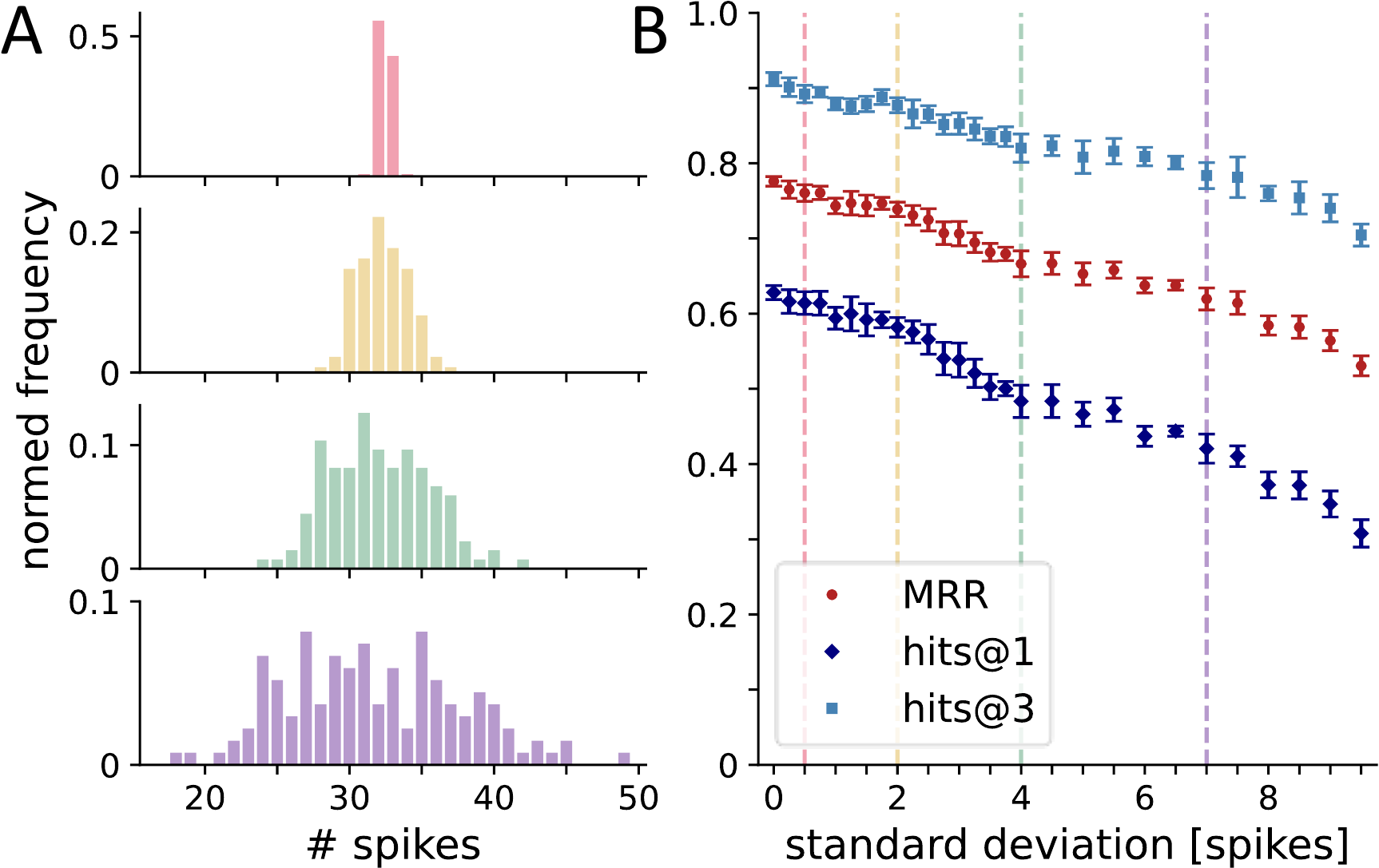}
	\caption{
	\textbf{(A)} Instead of a fixed length, spike train embeddings with variable length are used, with (from top to bottom) increasing standard deviation.
	\textbf{(B)} MRR and hits@k for spike train embeddings with variable length.
	The cases shown in (A) are marked with vertical lines.
	}\vspace{-4mm}	
	\label{fig:unequal}
\end{figure}

As shown in \cref{fig:unequal}, if the number of spikes per neuron varies only slightly (red, yellow), the obtained performance is almost unaffected.
For larger variations, the performance slowly drops (green, violet).
Therefore, the requirement that all embedding spike trains have the same length can be relaxed without impairing performance drastically.

\section{Neuronal implementation}\label{sec:neuronal}
The proposed coding scheme can be realized with arbitrary spiking neuron models, as long as gradients with respect to the output spike times can be calculated.
Currently, two widely applied options exist: either use a neuron model where the TTFS can be calculated analytically \citep{mostafa2017supervised,comsa2019temporal,kheradpisheh2019s4nn,goltz2021fast} or use surrogate gradients \citep{neftci2019surrogate}.
Here, we adopt the former approach and use a TTFS model to obtain interspike intervals.

\begin{figure}[t]
    \centering
    \includegraphics[width=0.975\columnwidth, page=3]{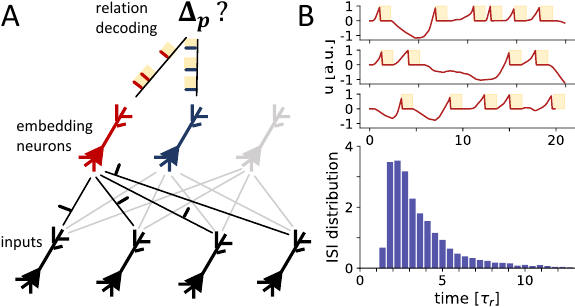}
	\caption{
	\textbf{(A)} Network architecture for learning spike train embeddings with integrate-and-fire neurons.
	\textbf{(B)} Example membrane traces of embedding neurons (top) and the ISI distribution over all embedding neurons (bottom).
	}\vspace{-4mm}	
	\label{fig:nlif}
\end{figure}

As a proof of concept, we represent a node $s$ in the KG by the spike train $\stime{s} \in \mathbb{R}^N$ of an integrate-and-fire neuron with exponential synaptic kernel $\kappa(x,y) = \heaviside{x-y} \exp\left(-\frac{x-y}{\taus}\right)$,
\begin{equation}\label{eq:dotu}
    \frac{\mathrm{d}}{\mathrm{d}t} u_{s,i}(t) = \frac{1}{\taus}\sum_{j} w_{s,ij} \, \kappa(t, t^\mathrm{I}_j) \,,
\end{equation}
where $u_{s,i}$ is the membrane potential of the neuron after its $i$-th spike, $\taus$ the synaptic time constant and $\heaviside{\cdot}$ the Heaviside function.
$w_{s,ij}$ are synaptic weights from a pre-synaptic neuron population, with every neuron $j$ emitting a single spike at fixed time $t^\mathrm{I}_j$ (\cref{fig:nlif}A).

After the neuron emits the $i$-th spike, it stays refractory for a time period $\taur$ and its membrane potential is reset to $0$, $u_{s,i+1}(0) = 0$. 
Afterwards, \cref{eq:dotu} is used to calculate the time-to-spike interval $I_{s,i+1}$, which is given by the time period needed for the membrane potential to cross the threshold $\thresh$ again, $u_{s,i+1}(I_{s,i+1}) = \thresh$.
The final spike train $\stime{s}$ is then obtained by summing up the obtained time-to-spike intervals and intermediate refractory periods
\begin{equation}
    t_{s,i} = \sum_{j\leq i} I_{s,j} + i \cdot \taur
\end{equation}

Given a KG $\mathcal{KG}\subset \mathcal{E} \times \mathcal{R} \times \mathcal{E}$, suitable spike trains are found by minimizing the margin-based ranking loss \cref{eq:loss} with respect to the weights $w_{s,ij}$ and the relation embeddings $\pmb{\Delta}_p$, $\forall s \in \mathcal{E}$ and $\forall p \in \mathcal{R}$.
For simplicity, each spike is caused by a disjunct subset of the input neurons with disjunct weights $w_{s,ij}$, avoiding unnecessary cross-dependencies during training\footnote{In principle, shared weights can be used for all spike times of a neuron as well, although we observed that this can complicate learning.}. 
However, the whole input population itself is shared by all node embedding neurons.

We apply the proposed spike-based graph embedding model to the IAD, UMLS and Kinships data sets (\cref{tab:mrr_nlif}), reaching similar performance levels as reported in \cref{tab:mrr}.
Our results further resemble those obtained for integrate-and-fire neurons using SpikE coding, which are reported as $\mathrm{MRR} = 0.78$ for the UMLS data set \citep{chian2021learning} and $\mathrm{MRR} = 0.65$ for the IAD data set \citep{dold2021spikeembed}.
For illustration, the membrane traces of three embedding neurons are shown in \cref{fig:nlif}B.
The interspike interval (ISI) distribution of the learned embedding spike trains resembles those seen for cortical spiking models \citep{nawrot2008measurement,ostojic2011interspike}. 
In addition, the (mean-corrected)\footnote{For the values chosen here, without mean correction, we get $\mathrm{CV} = 0.56$.} coefficient of variation -- given by $\mathrm{CV} = \frac{\sigma_\mathrm{ISI}}{\mu_\mathrm{ISI}-\taur}$, where $\sigma_\mathrm{ISI}$ and $\mu_\mathrm{ISI}$ are the standard deviation and 
\begin{table}[hbp]
\caption{Test performance with integrate-and-fire neurons.}
\label{tab:mrr_nlif}
\begin{center}\renewcommand{\arraystretch}{1.2}
\begin{tabular}{c | c c c}
\hline\hline
Data set & MRR & hits@1 & hits@3 \\
\hline\hline
IAD & 0.65 & 0.61 & 0.67 \\
\hline
UMLS  & 0.78  & 0.64 & 0.91 \\
\hline
Kinships & 0.43 & 0.26 & 0.50 \\
\hline \hline
\end{tabular}
\label{tab1}
\end{center}
\end{table}
mean of the ISI distribution, respectively -- is $\mathrm{CV} = 0.77$, lying in the regime of cortical spiking neurons which typically take values $> 0.5$ \citep{feng1999coefficient,nawrot2008measurement}.
Thus, the learned spike trains are similar to those usually encountered in functional spiking neural networks.
Details to all simulations can be found at the end of the document.

\section{Discussion}\label{sec:discussion}
Machine learning on KGs has recently experienced an increase in interest due to its applicability in many different areas, such as particle dynamics \citep{bapst2020unveiling}, material design \citep{xie2018crystal,rossusing}, question answering \citep{hildebrandt2020scene}, system monitoring \citep{soler2021graph}, industrial project design \citep{hildebrandt2018configuration,hildebrandt2019recommender} and spacecraft design \citep{berquand2019artificial}.
However, so far these methods have been mostly limited to classical machine learning algorithms and artificial neural networks.
Here, we introduce a graph embedding model based on spike trains, opening up the possibility of merging current SNN models with relational representation learning, e.g., to integrate different data modalities in a semantically meaningful way and perform relational reasoning tasks on them.
Both the abstract model (\cref{sec:current}) as well as its neuronal implementation (\cref{sec:neuronal}) are differentiable and can therefore be combined and trained end-to-end with subsequent SNN architectures, such as hierarchical and recurrent networks, equipping them with a learnable memory storing semantic information.
Different from previous embedding schemes \citep{dold2021spikeembed,chian2021learning}, each element in the graph is represented by the spike train of one corresponding neuron, fully utilizing the temporal domain of spikes to encode the relational structure of KGs.
The key aspect for moving from single spike to spike train embeddings is to formulate the gradient-based optimization with spike time intervals.

Compared to alternative approaches \citep{crawford2016biologically} based on semantic pointers, our model enables us to learn purely spike time-based and low-dimensional (i.e., low number of spikes per neuron) embeddings of graph elements.
This way, semantically meaningful and scalable spike representations can be flexibly learned for various settings, i.e., KGs with different link structure, number of relations and number of entities.
It is important to notice that KGs are usually very sparse, so only a tiny fraction of all possible triples are present in the data.
Therefore, they are often incomplete, meaning that not all true statements are contained in the KG but have to be inferred by the machine learning model.
KGs might even contain erroneous links, which the model has to detect from the structure of the KG during training. 
Graph embedding approaches have been demonstrated to work well in this setting \citep{nickel2015review,hamilton2017representation,ruffinelli2019you} by learning expressive embeddings that can subsequently be used in a variety of inference tasks, such as node classification (\cref{sec:detection}) and link prediction (\cref{sec:linkprediction}).

A potential disadvantage of the presented model (especially for realizations in physical substrates, such as analogue neuromorphic hardware \citep{billaudelle2019versatile}) is the dependence of the relation encoding on the $i$-th spike times of the node embedding neurons.
When faced with the possibility of unreliable spiking, where spikes can be dropped between trials, this correspondence breaks and scoring of triples becomes unreliable.
For instance, if the $i$-th spike time of population $s$ is missing, $t_{s,i+1}$ instead of $t_{s,i}$ will be compared with $t_{o,i}$ to evaluate the validity of a triple $(s,p,o)$.
However, we are confident that alternative and more expressive scoring functions can be found which are also compatible with spike-based coding and are more robust against spike drops -- similarly to how many different approaches for embedding knowledge graphs exist using real or complex-valued vector spaces in the machine learning literature \citep{ruffinelli2019you}.

To conclude, the presented results contribute to a growing body of research addressing the open question of how information can be encoded in the temporal domain of spikes.
In contrast to rate-based algorithms, purely spike time-dependent methods have the potential of enabling highly energy efficient realizations of artificial intelligence algorithms for edge applications, such as autonomous decision making and data processing in uncrewed spacecraft \citep{kucik2021investigating}.
Thus, by making novel data structures, methods and applications accessible to SNNs, the presented results are especially interesting for upcoming neuromorphic technologies \citep{wunderlich2019demonstrating,akopyan2015truenorth,mayr2019spinnaker,billaudelle2019versatile,pei2019towards,frenkel202028,orchard2021efficient,frenkel2021bottom}, unlocking the potential of joining event-based computing and relational data to build powerful and energy efficient artificial intelligence applications and neuro-symbolic reasoning systems. 

\section*{Acknowledgment}\label{sec:ack}
\addcontentsline{toc}{section}{Acknowledgment}
The author thanks Alexander Hadjiivanov and Gabriele Meoni for useful discussions and feedback on the paper. 
He further thanks his colleagues at ESA's Advanced Concepts Team for their ongoing support.
This research made use of the open source libraries PyKEEN \citep{ali2020benchmarking}, libKGE \citep{ruffinelli2019you}, PyTorch \citep{NEURIPS2019_9015}, Matplotlib \citep{Hunter:2007}, NumPy \citep{harris2020array} and SciPy \citep{2020SciPy-NMeth}.

\addcontentsline{toc}{section}{Simulation details}
\section*{Simulation details}

Simulations were done using Python 3.8.8 \citep{van1991interactively}, Torch 1.10.0 \citep{NEURIPS2019_9015} and PyKEEN (Python KnowlEdge EmbeddiNgs) 1.6.0 \citep{ali2020benchmarking}.

We use the following abbreviations for hyperparameters:
\begin{itemize}
    \item \textbf{dim}: embedding dimension,
    \item \textbf{LR}: learning rate,
    \item \textbf{L2}: L2 regularization strength,
    \item \textbf{BS}: batch size,
    \item \textbf{NS}: number of negative samples (generated randomly from the training graph) used during training,
    \item $\pmb \taur$: refractory period,
    \item $\pmb \gamma$: margin hyperparameter in the loss function,
    \item \textbf{IS}: size of input population stimulating the integrate-and-fire neurons.
\end{itemize}
Unless stated otherwise, L2 regularization strength and $\gamma$ are set to $0$.
The hyperparameters for the different models are summarized in \cref{app:tab1,app:tab2,app:tab3,app:tab4}.

\begin{table}[h]\renewcommand{\arraystretch}{1.2}
\caption{Hyperparameters for the experiments with the proposed spike-train embedding framework.}\label{app:tab1}
\center
\begin{tabular}{ c|  c|  c | c | c | c | c}
\hline\hline
Data set & dim & LR & BS & NS & $\taur$ & $\gamma$ \\
\hline
FB15k-237 & 128 & 0.001 & 4000 & 1 & 0 & 1\\
IAD & 12 & 0.1 & 100 & 2 & 0.08 & 0\\
UMLS & 32 & 0.01 & 100 & 10 & 0.03 & 0\\
Kinships & 32 & 0.01 & 400 & 4 & 0.03 & 0\\
Zachary Karate Club & 10 & 0.01 & 32 & 10 & 0.1 & 0 \\
\hline\hline
\end{tabular}
\end{table}

\begin{table}[h]\renewcommand{\arraystretch}{1.2}
\caption{Hyperparameters for the experiments with TransE.}\label{app:tab2}
\center
\begin{tabular}{ c|  c|  c | c | c | c | c}
\hline\hline
Data set & dim & LR & BS & NS & L2 & $\gamma$ \\
\hline
FB15k-237 & 128 & 0.001 & 4000 & 1 & 0 & 1\\
IAD & 12 & 0.1 & 100 & 2 & 0.0001 & 0\\
UMLS & 32 & 0.01 & 100 & 10 & 0 & 0\\
Kinships & 32 & 0.01 & 400 & 2 & 0 & 0\\
\hline\hline
\end{tabular}
\end{table}
\begin{table}[h]\renewcommand{\arraystretch}{1.2}
\caption{Hyperparameters for the experiments with RESCAL.}\label{app:tab3}
\center
\begin{tabular}{ c|  c|  c | c | c | c }
\hline\hline
Data set & dim & LR & BS & NS & L2 \\
\hline
FB15k-237 & 128 & 0.001 & 4000 & 1 & 0 \\
IAD & 12 & 0.1 & 100 & 2 & 0.00005\\
UMLS & 32 & 0.01 & 100 & 2 & 0.0001\\
Kinships & 64 & 0.01 & 400 & 2 & 0.0001\\
\hline\hline
\end{tabular}
\end{table}
\begin{table}[h]\renewcommand{\arraystretch}{1.2}
\caption{Hyperparameters for the experiments with integrate-and-fire neurons.}\label{app:tab4}
\center
\begin{tabular}{ c|  c|  c | c | c | c }
\hline\hline
Data set & dim & IS & LR & BS & NS \\
\hline
IAD & 12 & 50 & 0.1 & 100 & 2\\
UMLS & 64 & 50 & 1.0 & 100 & 10 \\
Kinships & 32 & 100 & 0.1 & 400 & 2 \\
\hline\hline
\end{tabular}
\end{table}

Gradient updates were applied using the Adam optimizer.
For TransE, we used the implementation of PyKEEN, while for RESCAL the PyKEEN implementation was only used for FB15k-237 and otherwise a custom implementation in Torch was used.
For CoDEx-S, we used the implementations with default parameters of LibKGE \citep{ruffinelli2019you}, where we created a custom implementation of SpikTE within LibKGE and used the TransE hyperparameters for training.

For the IAD data set, we used a soft margin loss for both TransE and SpikTE.
For RESCAL, we used a binary cross-entropy loss (on logits) for FB15k-237 and a mean squared error loss otherwise.

For the integrate-and-fire neurons, we used a threshold $\thresh = 1$, membrane time constant $\taum = 0.5$, refractory period $\taur = 0.1$ and total time interval $T=1$ for input spikes.
Presynaptic weights from the input population to embedding neurons were initialized from a normal distribution $\mathcal{N}\left(0.2, 1.0\right)$.
Similar to \citep{mostafa2017supervised,dold2021spikeembed,chian2021learning}, we added an additional regularization term to the cost to guarantee that neurons spike 
\begin{equation}
    L_\delta = 
\begin{cases}
    \sum_{s,i} \delta  \cdot \left(\thresh - W_{s,i} \right)     & \text{if } W_{s,i} \leq \thresh \,, \\
    0              & \text{otherwise}\,,
\end{cases}
\end{equation}
with $W_{s,i} = \sum_{j} w_{s,ij}$ and $\delta = 0.01\,$.

\printbibliography
\addcontentsline{toc}{section}{References}

\end{document}